\documentclass[letterpaper, 10 pt, conference]{ieeeconf}  % Comment this line out if you need a4paper

\IEEEoverridecommandlockouts                              % This command is only needed if 
\usepackage{graphicx}

\usepackage[caption=false]{subfig}
\usepackage{xcolor}
\usepackage{array}
\usepackage{float}
\usepackage{booktabs}
\usepackage{adjustbox}
\usepackage[noadjust]{cite}
\usepackage{multirow}
\usepackage{todonotes}
\usepackage{marginnote}

\usepackage[a-1b]{pdfx}

\title{\LARGE \bf
A Bidirectional Fabric-based Pneumatic Actuator for the Infant Shoulder: Design and Comparative Kinematic Analysis}

\author{Ipsita Sahin,$^{1}$ Jared Dube,$^{1}$ Caio Mucchiani,$^{2}$ Konstantinos Karydis,$^{2}$ and Elena Kokkoni$^{1}$% <-this % stops a space
\thanks{$^{1}$Dept. of Bioengineering; $^{2}$Dept. of Electrical and Computer Engineering, University of California, Riverside, 900 University Ave, Riverside, CA 92521, USA. Email:{\tt\footnotesize\{isahi001, jdube004, caiocesr, karydis, elenak\}@ucr.edu}. 
We gratefully acknowledge the support of NSF \# CMMI-2133084. 
Any opinions, findings, and conclusions or recommendations expressed in this material are those of the authors and do not necessarily reflect the views of NSF.
}}

\begin{document}
\maketitle
\thispagestyle{empty}
\pagestyle{empty}

%%%%%%%%%%%%%%%%%%%%%%%%%%%%%%%%%%%%%%%%%%%%%%%%%%%%%%%%%%%%%%%%%%%%%%%%%%%%%%%%
\begin{abstract}
This paper presents the design and assessment of a fabric-based soft pneumatic actuator with low pressurization requirements for actuation making it suitable for upper extremity assistive devices for infants. 
%with low pressurization for actuation, which can be used in devices to assist infants with arm movement.
%of which low pressurization requirements for actuation makes it suitable for assistive devices for infants.  
The goal is to support shoulder abduction and adduction without prohibiting motion in other planes or obstructing elbow joint motion. 
First, the performance of a family of actuator designs with internal air cells is explored via simulation. 
The actuators are parameterized by the number of cells and their width. Physically viable actuator variants identified through the simulation are further tested via hardware experiments. 
Two designs are selected and tested on a custom-built physical model based on an infant's body anthropometrics. 
Comparisons between force exerted to lift the arm, movement smoothness, path length and maximum shoulder angle reached inform which design is better suited for its use as an actuator for pediatric wearable assistive devices, along with other insights for future work.

%the designs were selected and results indicate better stability and smoothness of the two cell actuator between the performance of two designs analyzed

%in stability, smoothness and  of a two-cell actuator design . 

%is performing the best, and opened new dimensions for future work.  
%\textcolor{red}{Extend by including a summary of what is being developed and what are the key results.}
%
\end{abstract}

%\vspace{-3pt}
%%%%%%%%%%%%%%%%%%%%%%%%%%%%%%%%%%%%%%%%%%%%%%%%%%%%%%%%%%%%%%%%%%%%%%%%%%%%%%%%
\section{Introduction}

Though typically developing infants move their arms and explore their environment from the first months of age, infants with or at risk for upper extremity (UE) impairments show delayed emergence of UE motor milestones, and/or diminished ability to perform motor actions~\cite{Lobo2015CharacterizationInfancy}.
%add reference: Zablotsky2019
%Early intervention maximizes the opportunities for both the motor and cognitive learning~\cite{EE}. 
Providing infants with opportunities for training through the use of assistive devices can lead to improvements on their motor function \cite{Henderson2008}.
Currently, UE assistive devices for infants are limited~\cite{Arnold2020ExploringReview.} and are mainly passive (notable examples are the Pediatric Wilmington Robotic Exoskeleton, a rigid 3D-printed, two-link device~\cite{Babik2016}; and, PlaySkin Lift, a soft wearable with mechanical inserts~\cite{Lobo2016PlaySkin}).
Actuated devices, however, can deliver targeted and on-demand assistive forces that could directly adapt to the user as they grow and their abilities change~\cite{Sugar2007} (e.g., compared to the need for manual adjustment of mechanical assistance, as seen in~\cite{Babik2016, Lobo2016PlaySkin}).
Our recent work on the development of an actuated UE wearable device for infants has utilized silicone-based pneumatically-controlled actuators to provide assistance at both elbow and shoulder joints~\cite{Kokkoni2020_asme}. 
The current paper extends this work by introducing new soft actuator designs for the shoulder joint.

Soft materials and compliant structures are increasingly being used for actuation in assistive and rehabilitation devices~\cite{Polygerinos2013,Maeder-York2014, Polygerinos2015, Park2014,Yap2017,Nguyen2019}.
%removed Low2018
To be functional and practical, soft robotic actuators need to meet certain design specifications, such as high power density, accuracy and repeatability in both manufacturing and performance~\cite{Laschi2013, Iida2011, Agarwal2016, Majidi2014}. %, Baniqued2016 consider this?
%which can be, at times, conflicting. 
%DONE \textcolor{red}{add a few references to backup this statement}. 
% \textcolor{red}{This needs significant editing.  I think you want to say about soft pneumatic actuators in general (which I agree), but EIAs is not the only type. List other designs too, ans also start by saying why pneumatics and not cable-driven or other types. Also, you should start by saying what are the design specifications for soft actuators (I fixed this now via the previous sentence, but keep in mind for future).}
%Soft wearable devices are considered safer to wear and use, as well as more comfortable and aesthetic, without sacrificing functionality or performance.
%This calls for the development of assistive and rehabilitation technologies for infants with or at risk of UE motor impairments.
%For example, although cable-driven devices are good for providing a great amount of assistive force, they may lag inherent safety and )~\cite{Xiong2020}. 
Soft pneumatic actuators are lightweight, have simpler control specifications, %[ALTHOUGH SOFT ACTUATORS'S CONTROL IS CHALLENGING]}, 
and provide inherent safety (i.e. low risk of getting injured in case of malfunction) over other complex engineered devices~\cite{Xiong2020}. 
Commonly used soft pneumatic actuators are Elastic Inflatable Actuators (EIAs), which include silicone-based EIAs and fabric-based EIAs, among others.
%PneuNets, and Pneumatic Artificial Muscles (PAM).  
%There are advantages and disadvantages to these designs.

EIAs are powered by pressurized gas (or liquid) and are capable of achieving large strokes with very little friction, thus exhibiting distributed force generation~\cite{Gorissen2017}. 
%because of no sliding part
%due to Pascal’s principle
Compared to silicone-based EIAs, fabric-based EIAs require lower pressure because of the non-extensibility of the fabric material.
Also, fabric-based EIAs can be built faster and at a lower overall cost than silicone-based ones. 
Fabrication of the latter involves several steps that may take hours to complete: 3D printing molds for different parts (e.g., $\sim$1.5 hrs~\cite{Yang2017}, etc.), multi-staged casting and curing of silicone
%\footnote{https://www.smooth-on.com/product-line/dragon-skin/}
(e.g., 0.5-16 hrs for DragonSkin, etc.), and assembling parts using suitable adhesives. 
%(e.g., Dragon Skin\textsuperscript{TM}) 
%Hence, it can take more than half a day to a few days to fabricate a silicone-based actuator, sometimes even more depending on the complexity of the design. 
% The lighter the actuator, the lesser inertia resulting in lower contact force and lower overall momentum which is important to maintain safety.
% On the other hand, EM actuators are heavy, add mass and inertia to the system and are of concern with the safety measures.
%The fabric-based actuators fulfill the design compatabilities.
On the other hand, fabric-based soft pneumatic actuators, such as those made of flexible Thermoplastic PolyUrethane (TPU) coated nylon fabric, can be built fast, in a cost-effective manner, and without the use of any specialized equipment \cite{Low2017}.
Even fabric-based actuators with complicated designs (e.g., use of pleats on the actuator to guide motion during inflation~\cite{Sanan2014}, application of double-layered inner bladders to prevent air-leakage~\cite{Best2015}, fabrication of wrinkle actuators using specialized equipment like high-frequency electromagnetic welding~\cite{Park2020},  etc.) that do require significant fabrication time, may still be made faster than their silicone counterparts with similar functionality.

Fabric-based shoulder actuators are seen in UE devices intended for use by adults~\cite{Simpson2017, Simpson2020}, and children~\cite{Li2019DesignExoskeleton} but not infants (exception is \cite{Paez2021} which contains underarm actuators for the purpose of assisting postural transitions from lying to sitting).
%shoulder abduction/adduction
In this work, we explored the use of fabric-based EIAs as an alternative to the silicon-based EIAs previously used for our application~\cite{Kokkoni2020_asme}.

\section{Fabrication Process, Design Considerations, and Preliminary Feasibility in Simulation}
A new family of flexible TPU fabric-based, multi-pouched actuators for shoulder abduction/adduction is presented. 
%The actuators are parameterized based on the number and width of air pouches (cells). 
Initially, first-principles-based numerical simulation was conducted to determine the preliminary feasibility of all proposed actuator variants prior to physical testing. 
Down-selected actuator variants were then fabricated and tested on a custom-built physical infant model. 
%Force exerted to lift the arm, end-effector movement smoothness and range of motion were evaluated.
%We begin with the fabrication process of our proposed actuators. Simulation is used to assess the feasibility of various actuator variants prior to physical testing. 
\subsection{Design Approach}
\begin{figure}[!t]
\vspace{6pt}
     \centering
     \includegraphics[width=0.9\columnwidth]{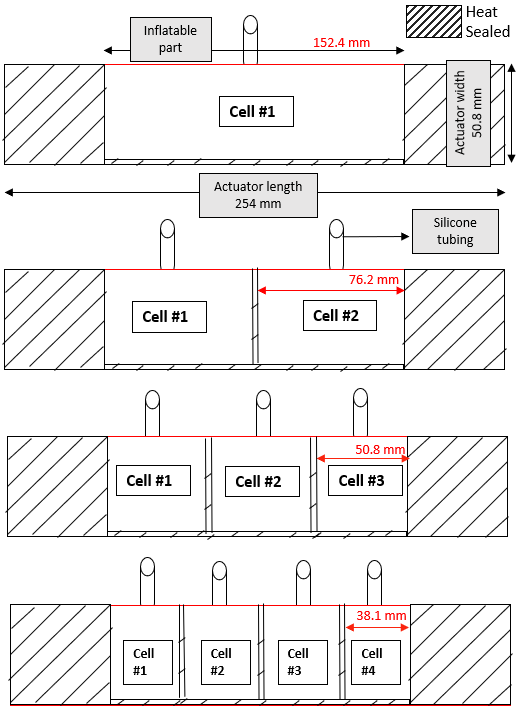}
     \vspace{-9pt}
         \caption{Cross-section view of the proposed family of parameterized soft fabric-based pneumatic actuators considered in this work.}
     \label{fig: 4 cells}
     \vspace{-12pt}
\end{figure}

The design of our proposed fabric-based actuators with air pouches (cells) (Fig.~\ref{fig: 4 cells}) improves upon the design of the silicone-based shoulder actuators in~\cite{Kokkoni2020_asme}. 
The new actuators are low-profile (i.e. non-obtrusive when not inflated) and can be fabricated fast ($<$20 min). We use flexible and lightweight fabric (Oxford 200D heat-sealable coated fabric [properties listed in Table~\ref{size}]). 
The fabric can be heat-sealed via non-specialized equipment like a household iron. 
To reduce areas that may become potential leakage points, we cut strips of fabric, wrap them around, and seal together the two opposite coated sides (instead of cutting two separate pieces and seal them at their periphery). 
This can constrain potential leakage to the point where tubing is embedded into a cell.  

\begin{table}[!ht]
\vspace{-6pt}
\caption{Nylon-Oxford Fabric Properties}\label{size}
\vspace{-12pt}
\begin{center}
\begin{tabular}{l l}
\toprule
Elastic Modulus $(N/m^2)$ & 498000000 \\
Poisson's Ratio $(N/A)$ & 0.35 \\
Shear Modulus $(N/m^2)$ & 184400000\\
Mass Density $(kg/m^3)$ & 757.58\\
Tensile Strength $(N/m^2)$ & 17520\\
Compressive Strength $(N/m^2)$ & 103421000\\
Yield Strength $(N/m^2)$ & 58605000\\
Thermal Expansion Coefficient $(1/K)$ & 1.00E-06\\
Thermal Conductivity $(W/(m-K))$ & 0.53\\
Specific Heat $(J/(kg-K))$ & 1386\\
\bottomrule
\end{tabular}
\end{center}
\vspace{-12pt}
\end{table}

We focused on the design of rectangular actuators (Fig.~\ref{fig: 4 cells}). 
The actuators contain non-inflatable and inflatable portions, and are parameterized based on three characteristics: i) their total length (kept herein constant at $254\;$mm), ii) their width (we consider five cases of $50.8\;$mm, $44.45\;$mm, $38.1\;$mm, $31.75\;$mm and $25.4\;$mm), and iii) the number of discrete cells the inflatable part is divided into ($1-4$ cells). 
%The actuator comprises non-inflatable and inflatable portions. 
The non-inflatable portion contains two $50.8\;$mm-long areas at each side, used for attaching the actuators onto the body.
The inflatable portion is the area in the middle and measures $152.4\;$mm in length.
Table~\ref{tab:simulation} contains the 20 distinct actuator variants considered in this work.\footnote{This type of design can directly scale up to create larger actuators, e.g., targeted to older children. However, scaling down is determined by the available fabrication means to heat-seal the fabric (smaller actuators require thinner heat-sealed seams which may be achieved e.g., by low-heat precision-tip ironing) as well as its practical utility given the application context. For instance, as discussed later, actuators with more than four cells do not generate enough force required for the present application.} 
Their main dimensions (importantly length and width) were informed by anthropometric data of the infant population reported in the literature. 
The upper arm length for male and female infants of 6-24 months of age varies on average between $142-188\;$mm and $135-183\;$mm, respectively~\cite{Fryar2021}. 
%The actuators aim to cover the upper arm- underarm- mid-axillary line. 

%We consider four different versions whereby the inflatable portion is divided into 1-4 equal parts (referred to in the remainder of the paper as cells). 
%That design required three days to make one actuator and enforced a non-zero angle between the torso and the arm even when not inflated because of the mass of the actuator. %
%The proposed fabric-based approach addresses both limitations while also significantly reducing the weight of the actuator. 

%to have inflation all over the actuator instantaneously.
%Keeping the length of the actuator constant, the width was varied as $50.8\;mm$, $44.45\;mm$, $38.1\;mm$, $31.75\;mm$ and $25.4\;mm$ for the 1 cell, 2 cell, 3 cell and 4 cell design, resulting into 20 different samples of the shoulder actuators. \footnote{\kkb{Note to self: mention scalability up and down here}}

\begin{table}[!t]
\vspace{6pt}
\caption{Key Actuator Design Parameters}\label{tab:simulation}
\vspace{-12pt}
\begin{center}
\begin{tabular}{c c c c}
\toprule
\multirow{1}{*}{Types} & Cell Length (mm) & Total length (mm) & Width (5 cases)\\
%\cmidrule[.25pt]{1-4}
\midrule
\multirow{1}{*}{1 Cell} & \multirow{1}{*}{152.40} &\multirow{1}{*}{254}  &  $\mathcal{W}$\\
%\cmidrule[.25pt]{1-4}
 \multirow{1}{*}{2 Cell} & \multirow{1}{*}{76.20} &\multirow{1}{*}{254} & $\mathcal{W}$\\
%\cmidrule[.25pt]{1-4}
\multirow{1}{*}{3 Cell} & \multirow{1}{*}{50.80} &\multirow{1}{*}{254} & $\mathcal{W}$\\
%\cline{1-4}
\multirow{1}{*}{4 Cell} & \multirow{1}{*}{38.10} &\multirow{1}{*}{254} & $\mathcal{W}$\\
\midrule
\multicolumn{4}{l}{* where {$\mathcal{W}=\{50.80, 44.45,38.10,31.75,25.40\}\;$mm}}\\
\bottomrule
\end{tabular}
\end{center}
\vspace{-18pt}
\end{table}

Our proposed fabric-based actuators are lightweight, making them ideal for use in pediatric assistive and rehabilitation devices. Specifically, their weight varies between $3.8\;$g for thinner ones and $8.2\;$g for wider ones, which is a reduction of over $84\%$ compared to the silicone-based ones~\cite{Kokkoni2020_asme} that weigh about $51.2\;$g each. 
Lightweight actuators result in smaller inertia and hence lower contact force and lower overall momentum, which is critical for safety.
%To understand the design feasibility and actuator's ability to perform, 20 of the actuator designs were simulated using SolidWorks (Dassault Systemes,Waltham, MA) and two of the designs (i.e. 1 cell and 2 cell $50.8\;mm$ width) were experimentally tested on a wooden model of infant. 

\subsection{Feasibility Assessment in Simulation}
The performance and design feasibility of the proposed family of actuator designs (Table~\ref{tab:simulation}) was first assessed through simulated experiments (SolidWorks; Dassault Systems, Waltham, MA). 
Simulations were used to estimate the stress, strain, deformation scale, and displacement, to identify actuator designs that can maximize the force output to be useful for arm abduction without failing.  Simulated experiments serve as the basis to guide hardware development in the sense of filtering out some of the parametrization listed in Table~\ref{tab:simulation} that are not expected to be physically viable. 

Simulations used the TPU-coated nylon from the software's material library with yield strength of $5.86e^{7}\;N/m^2$. 
All components of the simulated actuators employed the shelling function on $10\;mm$ thick cells to create $9\;mm$ pockets within all the given actuators cells, which closely matches the height of the physical actuators. 
An outward pressure from within the cells against their inner surfaces was applied to find the maximum pressure up to which the actuators avoid functional failure.

\begin{figure}[!t]
\vspace{6pt}
     \centering
     \includegraphics[width=\columnwidth]{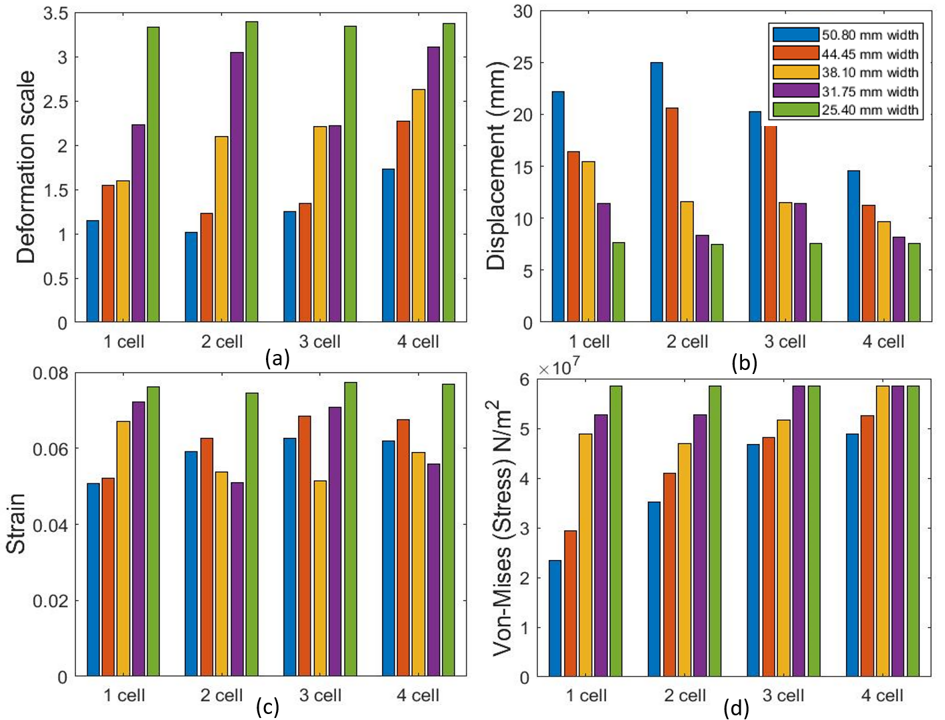}
     \vspace{-18pt}
         \caption{Effects of decreasing width for all $20$ actuators. %s for Top: (a) Deformation scale, (b) Displacement; Bottom: (c) Strain, (d) Stress. 
         Deformation scale, strain and stress increase but displacement decreases with smaller widths.}
     \label{fig: solidwork}
\end{figure}

\begin{figure}[!t]
\vspace{-6pt}
     \centering
     \includegraphics[width=\columnwidth]{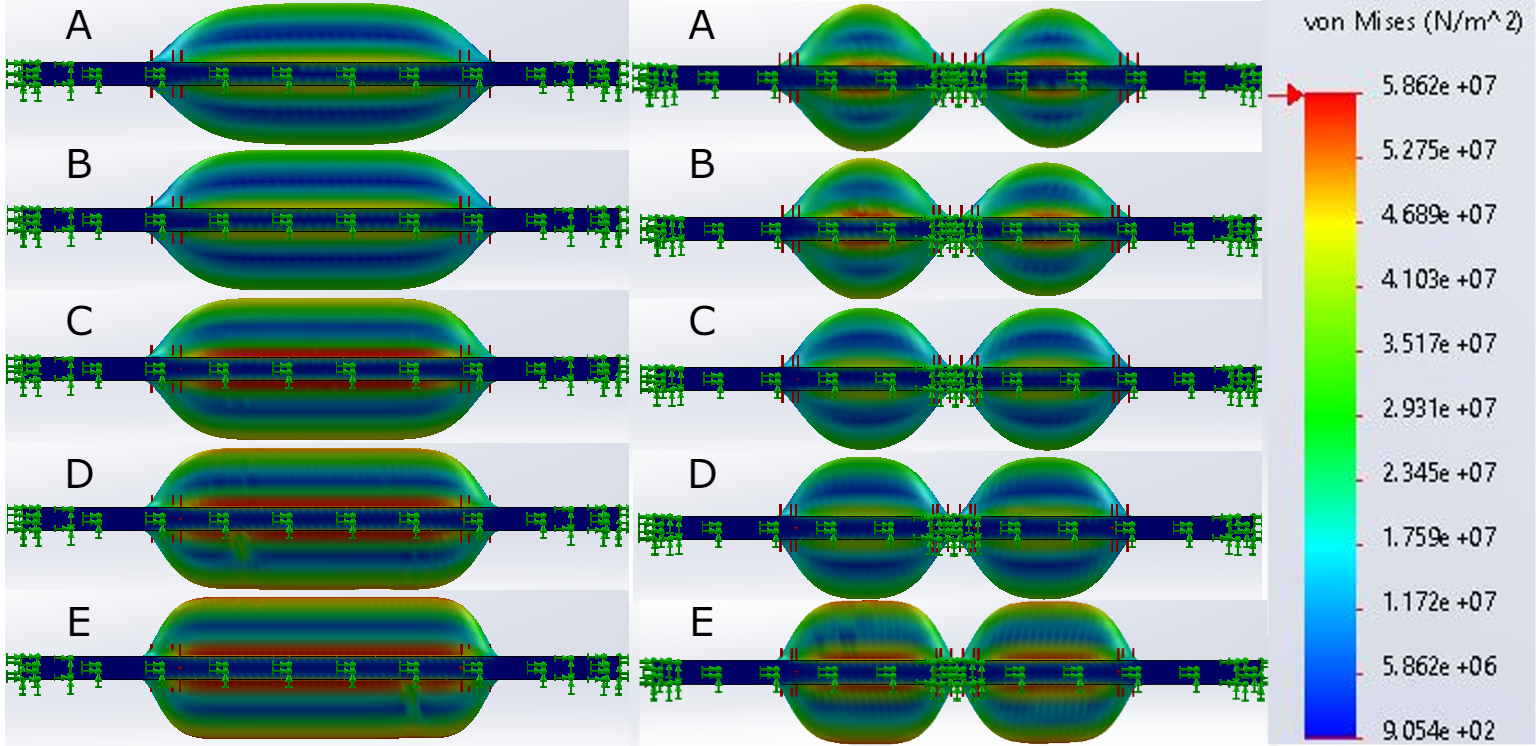}
     \vspace{-18pt}
         \caption{Simulation for the shoulder actuators for various geometries i.e. decreasing the width from A (50.80mm) to E (38.10mm) ($top-bottom$) of Von Mises stress contour plots for stresses in 1 cell ($left$) and 2 cell ($right$). As the width decreases, the von Mises value increases.}
     \label{fig:5cell}
     \vspace{-18pt}
\end{figure}

%The simulations enables us to make useful observations to further guide hardware design and testing. 

It can be seen in Fig.~\ref{fig: solidwork} the cell width is directly proportional to the amount of displacement, whereas the deformation scale and strain are inversely proportional. %(i.e. if the cell width decreases, the displacement decreases too but the strain and the deformation scale increases).
% This supports that the actuators will perform the best at a higher cell size, thus the 1 and 2 cell with the highest width value become the best choice for the abduction of the arm with low possible chance of failure. 
Furthermore, a decrease in cell width (from $50.8\;$mm to $25.4\;$mm) causes a critical error as von Mises values exceed the yield strength and strain is much higher (Fig.~\ref{fig:5cell}). 
To determine the optimal number of cells, note the increasing trend among number of cells, von Mises and strain. This suggests that actuators with fewer cells and higher width are expected to perform better. 

Armed with this analysis, 1- and 2-cell actuators of the highest width ($50.80\;$mm) were deemed to be most viable choices for shoulder abduction/adduction with lower chances of failure, and hence were also tested experimentally. 

%Based on these observations, the physical experiment focused on the 1 cell and 2 cell actuators ($50.80\;mm$ width) and not the remaining eighteen designs.

% Then in regards to the amount of cells you will see the same trend as increasing the number of cells from one to 4, the Von-Mises will increase, strain will increase, deformation scale, and displacement will decrease.
% Using this evidence, we reasoned that it was not essential to physical experimental trials for the 3 and 4 cells designs but instead focused on testing only the 1 and 2 cell actuators for our device.
% The 3rd trial with 1mm width ends up being a critical failure for each of the 4 designs so they 3rd data point for each design was excluded from figure 1 as to not skew the graph.
%Based on these observations, the physical experiment focused on the 1 cell and 2 cell actuators ($50.80\;mm$ width) and not the remaining eighteen designs.
% The 3rd trial with 1mm width ends up being a critical failure for each of the 4 designs so they 3rd data point for each design was excluded from figure 1 as to not skew the graph.

\section{Experimental Testing and Evaluation}

%\subsection{Actuation and Pneumatic Control}
\subsection{Experimental Setup}
\begin{figure}[!t]
\vspace{6pt}
\centering
\subfloat[The experimental setup with the actuator placed under the arm.]{%
  \includegraphics[width=0.9\columnwidth]{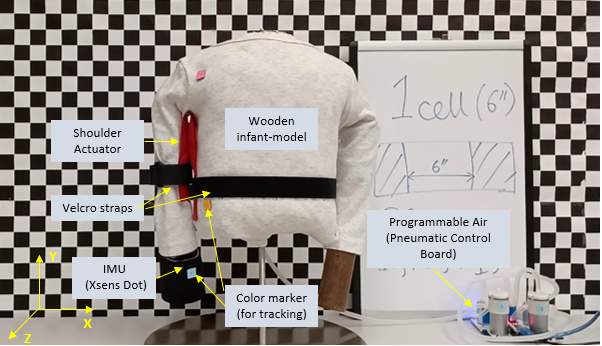}%
}
\vspace{-6pt}
\subfloat[1-cell (left) and 2-cell (right) physical actuator prototypes at deflated state (top) and at full inflation (bottom).]{%
  \includegraphics[clip,width=0.95\columnwidth]{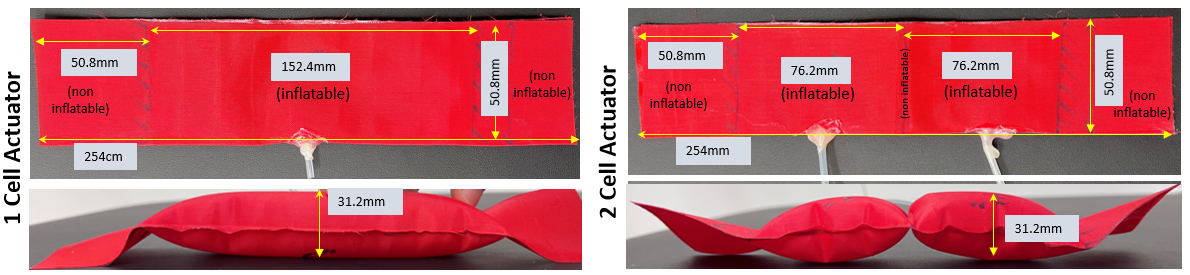}%
}
\caption{Physical experimentation setup and sample actuator prototypes.}\label{fig:setup}
\vspace{-24pt}
\end{figure}

%While the actuator is the on-suit device for the wearable, the 
Inflation and deflation of the actuators were controlled using an-off body pneumatic control board (Programmable-Air---an open-source hardware kit to control inflatable soft robots).  %\footnote{https://github.com/Programmable-Air} 
The board has two compressor/vacuum pumps and three pneumatic valves to control the airflow during inflation and deflation and weighs $0.35\;$kg (Fig.~\ref{fig:setup}a). The pumps have a flow rate of $2\;$lit/min. The Pulse Width Modulation (PWM) or the duty cycle of the pump can be varied from $0\%$ to $100\%$, and although the pump may turn on around $20$\% duty cycle, the lower the $\%$ duty cycle, the longer it will take to inflate/deflate. The board can create pressure varying between $[-7.5, +7.5]\;$psi. 
An Arduino Nano (ATMega328P) connects the board to a computer via microUSB.
This electro-pneumatics board is powered up using a 12V adapter.
The board also has a pressure sensor (SMPP-03). 
After full inflation, if a certain internal pressure is reached, airflow automatically stops to prevent leakage and/or actuator damage. 

For the physical experiments, the two down-selected actuator variants (1-cell and 2-cell actuators; Fig.~\ref{fig:setup}b) were placed on a physical model scaled approximately to the 50\textsuperscript{th} percentile of a 12-month-old infant's upper body~\cite{Fryar2021} using velcro straps on the non-inflatable portions (Fig.~\ref{fig:setup}a).
%
%The infant model has a rigid torso, two rigid upper and forearm part, with each arm weighting $0.45\;kg$. 
%
%The forearms are connected to the upper arms through 1-DOF hinge-elbow joints, whereas 
The arm was linked to the torso through a 3-DOF ball-and-socket shoulder joint. 
%Thus each arm-torso subsystem has 4-DOFs. 
%During the experiment, the wearable actuator was actuated and controlled using a pneumatic control board (the Programmable-Air) to lift the arm of the wooden model against gravity i.e. the abduction and then again reach the home position during adduction. 
%During abduction, the center of gravity of the arm acts slightly above the elbow joint.
%It would be unobtrusive to any other movement of the infant when powered off.
Actuator placement was selected such that it can support shoulder abduction/adduction without hindering rotation about the other axes. 
\begin{figure*}[!t]
\vspace{6pt}
\begin{minipage}{0.5\textwidth}
\subfloat[Shoulder $abduction$ during inflation of the {\bf 1-cell} actuator.]{%
  \includegraphics[clip,width=0.925\columnwidth]{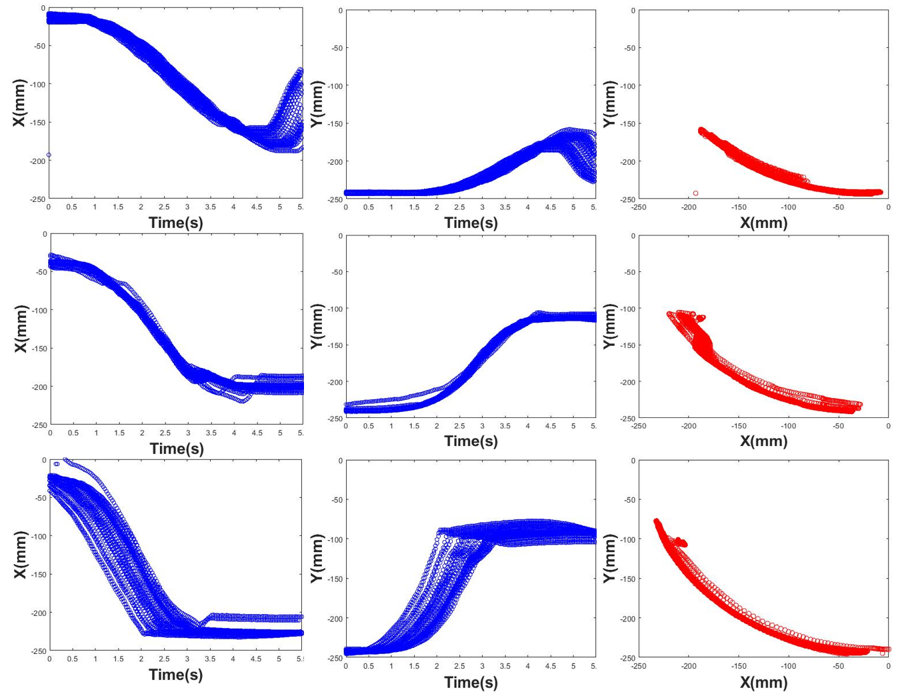}%
}
\vspace{-6pt}

\subfloat[Shoulder $adduction$ during deflation of the {\bf 1-cell} actuator.]{%
  \includegraphics[clip,width=0.925\columnwidth]{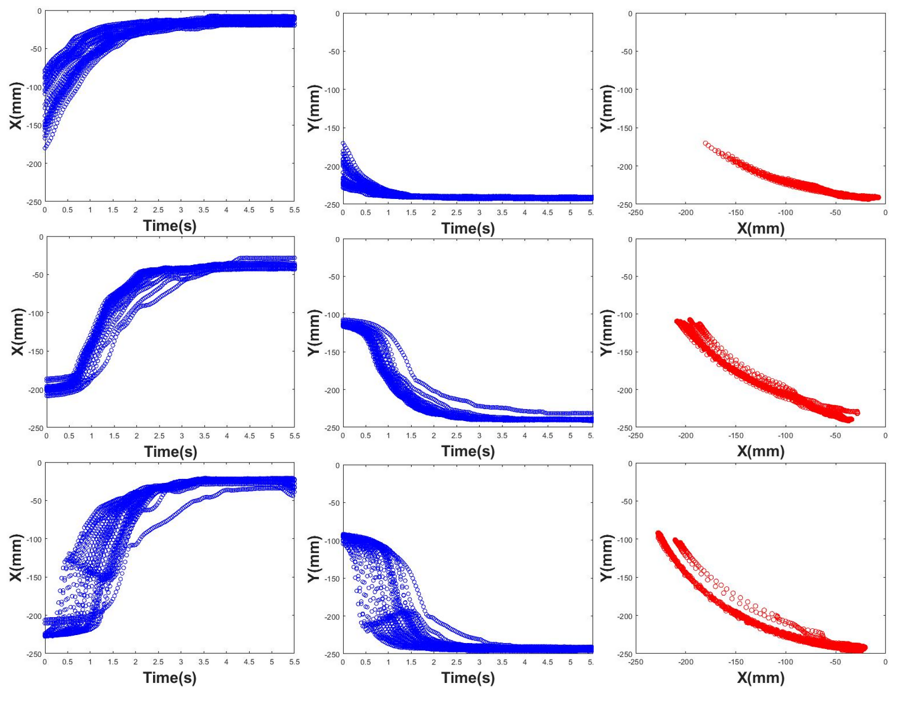}%
}
%\vspace{-12pt}
\end{minipage}
%\caption{Right arm's end-effector X and Y position, and the two-dimensional trajectories during abduction and adduction for the 1 cell actuator. Curves in the top, middle and bottom row corresponds to 50\%, 75\% and 100\% duty cycle.}\label{fig:1cell}
%\vspace{-18pt}
%\label{fig:xy1}
%\end{figure}
%
%\begin{figure}[!t]
%\vspace{-12pt}
\begin{minipage}{0.5\textwidth}
\subfloat[Shoulder $abduction$ during inflation of the {\bf 2-cell} actuator.]{%
  \includegraphics[clip,width=0.925\columnwidth]{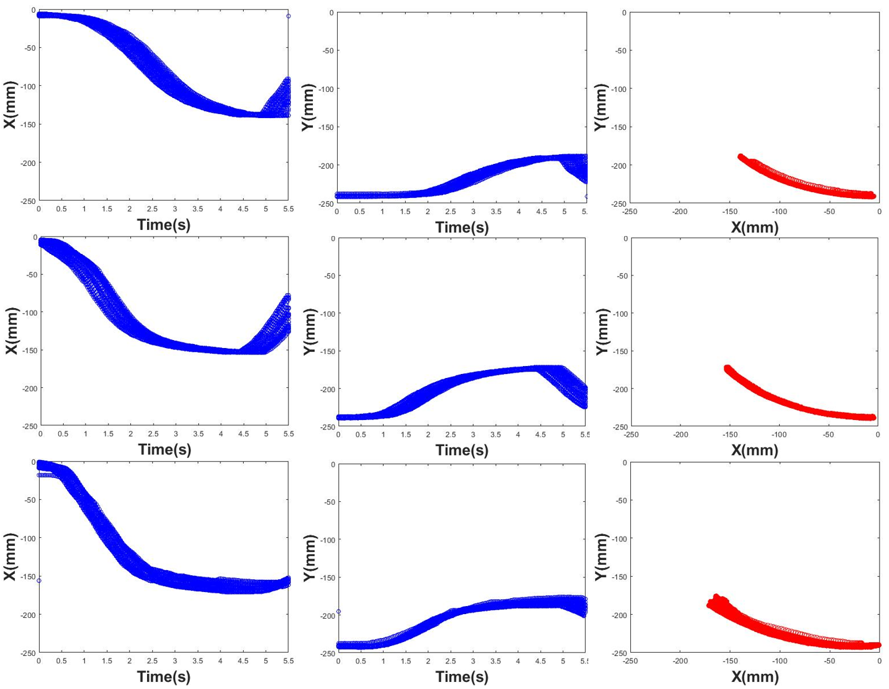}%
}
\vspace{-6pt}

\subfloat[Shoulder $adduction$ during deflation of the {\bf 2-cell} actuator.]{%
  \includegraphics[clip,width=0.925\columnwidth]{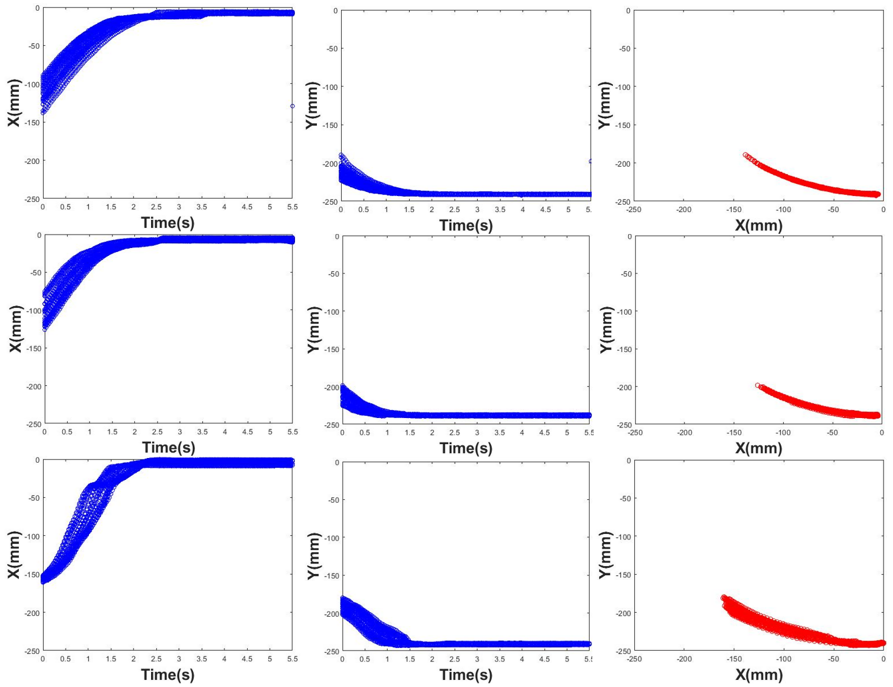}%
}
\end{minipage}
%\vspace{3pt}
\caption{Right arm's end-effector X and Y position, and the two-dimensional trajectories (curves in red) during abduction and adduction for the 1- and 2-cell actuators. Curves in the top, middle and bottom row correspond to $50\%$, $75\%$ and $100\%$ duty cycle, respectively.}\label{fig:2cell}
\vspace{-15pt}
\label{fig:xy2}
\end{figure*}
In the experiments, different conditions were considered by varying i) the duty cycle of the pneumatic control board ($50\%$, $75\%$ and $100\%$), and ii) the inflation/deflation times ($[1-5]\;$sec in $1\;$sec increments). 
Both actuators went through 30 trials of abduction and adduction for each condition.
Video recordings from the experiments were used to extract the 2D position of the end-effector (indicated by a color marker Fig.~\ref{fig:setup}(a)) of the arm (www.kinovea.org; v0.9.5).  %\footnote{www.kinovea.org}, 
%the 2D position of the end-effector (indicated by a color marker Fig: \ref{fig:setup}(a)) of the arm was annotated.
%
%@Ipsita: Describe the experiment first and then refer to the sensors that collected the data. Otherwise the reader will not be able to understand why you needed the sensors until they read further down. 
An Inertial Measurement Unit (IMU) (Xsens DOT; Xsens Technologies B.V., Enschede, The Netherlands) was placed at the end-effector to collect kinematic data (Euler angles, acceleration and velocity) at a sampling rate of $60\;$Hz. 
Lastly, the force exerted by the arm was measured with the actuator powered on, using a four-wire load cell and \textit{HX711} amplifier. 

To assess actuator performance we evaluated the end-effector's: i) exerted force, ii) movement smoothness (using Spectral Arc length [SPARC]~\cite{Melendez2021}), and iii) motion path length, as well as the maximum angle achieved at the shoulder joint during abduction/adduction. %, i.e. the maximum range of motion.

%We have followed the formula and the steps of SPARC analysis mentioned in \cite{SPARC1}.

\subsection{Experimental Results and Discussion}

To reach full inflation at $\{50, 75, 100\}\%$ duty cycle, it takes $\{4.5,3.5,2.7\}\;$sec and $\{3.5, 2.5,2\}\;$sec on average for the 1-cell and 2-cell actuators, respectively. This suggests a roughly superlinear decrease of inflation time as PWM increases, for both actuators.  Fig.~\ref{fig:xy2} depicts the 2D position data of the end-effector during shoulder abduction and adduction for the 1-cell and 2-cell, respectively. 

\subsubsection{Force Generation} 
\begin{figure}[!t]
\vspace{2pt}
     \centering
     \includegraphics[width=1\columnwidth]{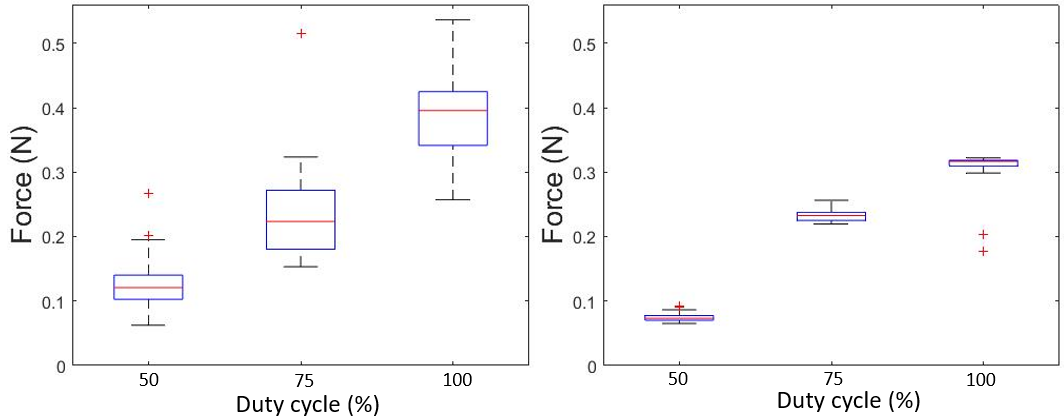}
     \vspace{-9pt}
     \caption{The force exerted by the arm during full inflation for 1-cell (left) and 2-cell (right) shoulder actuators at $50\%$, $75\%$ and $100\%$ duty cycle.}
     \label{fig:force}
     \vspace{-18pt}
\end{figure}

Forces exerted by the end-effector under 1-cell and 2-cell actuators are shown in Fig.~\ref{fig:force}.
A two-way Analysis of Variance (ANOVA) was performed to identify differences in force profiles between the two types of actuators and across the different duty cycles. 
A statistically significant interaction between number of cells and $\%$ duty cycle was found ($F[2, 180] = 12.36$, $p = 0.000$) supporting that the number of cells has an effect on force generation at the $50\%$ and $100\%$ duty cycles but not at the $75\%$. In the aforementioned two duty cycles, the 1-cell actuator is the one that produces greater forces. Overall, force generation increases with increasing $\%$ duty cycle.
%The force exerted by 1-cell actuator($M=0.25N$, $SD=0.12N$) is significantly greater ($F[1,180]=47.31$,$p=0$) than the 2-cell actuator($M=0.20N$, $SD=0.10N$), whereas at least two duty cycles significantly differ from each other ($F[2, 180] = 422.71$, $p = 0.0$). Tukey post-hoc test revealed that during 100\% duty cycle ($M=0.35N$, $SD=0.07N$) exerted force is significantly higher ($p=0.0$) than 75\% ($M=0.23N$, $SD=0.05N$), whereas force exerted by 75\% duty cycle is significantly higher ($p=0.0$) than 50\% ($M=0.10N$, $SD=0.04N$) duty cycle.
%The preliminary findings on the range of force provided by different duty cycles of different actuators may help to design a force-feedback closed loop controller to provide desired force during targeted reaching action.

% Multiple independent t-tests were performed to identify differences in force profiles between the two types of actuators and across the different duty cycles. 
% No significant difference between the cells at $75$\% valve duty cycle ($p=0.88$) is observed. 
% For the $50$\% and $100$\% duty cycles, the force exerted by the 1-cell actuator ($M\textsubscript{50\%}=0.13N$, $SD\textsubscript{50\%}=0.04N$; $M\textsubscript{100\%}=0.39N$, $SD\textsubscript{100\%}=0.07N$, ) is significantly greater ($p\textsubscript{50\%}=0$, $p\textsubscript{100\%}=0$) than the 2-cell one ($M\textsubscript{50\%}=0.07N$, $SD\textsubscript{50\%}=0.01N$; $M\textsubscript{100\%}=0.31N$, $SD\textsubscript{100\%}=0.03N$).

\subsubsection{Movement Smoothness}
Two-way ANOVAs were performed to identify differences on smoothness across the two types of actuators and the different duty cycles, for each type of motion (abduction and adduction).
For abduction, a significant interaction between number of cells and \% duty cycle was found ($F[2, 539] =37.50$, $p = 0.000$) supporting the \% duty cycle affects differently the performance of the two actuators.
For adduction, a significant difference only for the main effect of \% duty cycle was found ($F[2, 539] =85.08$, $p = 0.000$). Tukey post-hoc analysis showed that smoothness is significantly less at the 50\% duty cycle compared to the other two regardless of the number of cells ($p\textsubscript{50\%-75\%}=0.000$, $p\textsubscript{50\%-100\%}=0.000$, $p\textsubscript{75\%-100\%}=0.645$). 
Smoothness values along the axes (i.e. x, y and z) from IMU readings are shown in Table~\ref{smoothness}.\footnote{%~Estimating movement smoothness is very important to access the quality of the movement. 
~We used the continuous data streams of gyroscopic data collected in the IMU on a SPARC-based approach to quantitatively measure the movement smoothness~\cite{Melendez2021,Beck2018}. 
The adaptive cut-off frequency was determined to be $5\;Hz$ from the signal power spectral density. The signals were low-pass filtered with a zero-lag second order Butterworth filter with a cut-off frequency at $5\;Hz$ to remove the high-frequency noise.} 
%does not depend on the number of cells but only on the \%duty cycle.

\begin{table}[!h]
\vspace{6pt}
\caption{Smoothness (S) values}
\label{smoothness}
\vspace{-12pt}
\begin{center}
\begin{adjustbox}{width=\columnwidth,center}
\begin{tabular}{c c c c c c}
\toprule
\multirow{2}{*}{Movement} & \multirow{2}{*}{Sample} & \multirow{2}{*}{Duty cycle} & \multicolumn{3}{c}{Smoothness (S)(Mean$\pm$ SD)}  \\
\cmidrule[.25pt]{4-6}
& & & S\textsubscript{x}& S\textsubscript{y}& S\textsubscript{z} \\
\midrule
\multirow{6}{*}{Abduction} & \multirow{3}{*}{1 cell} & 50\% &$-7.67\pm1.99$ &$-3.85\pm1.61$ &$-2.4\pm0.50$\\
& & 75\% &$-4.56\pm0.63$ &$-5.34\pm0.84$ &$-1.85\pm0.08$\\
& & 100\% &$-3.27\pm0.76$ &$-2.08\pm0.30$ &$-1.97\pm0.07$ \\
\cmidrule[.25pt]{2-6}
%2 cell
& \multirow{3}{*}{2 cell} & 50\% &$-4.48\pm0.54$ &$-1.91\pm0.06$ &$-1.85\pm0.70$\\
& & 75\% &$-3.30\pm0.16$ &$-1.90\pm0.11$ &$-1.61\pm0.08$\\
& & 100\% &$-4.64\pm0.81$ &$-2.76\pm0.65$ &$-1.88\pm0.06$ \\
\midrule
%1 cell
\multirow{6}{*}{Adduction} & \multirow{3}{*}{1 cell} & 50\% &$-5.30\pm0.86$ &$-6.15\pm0.98$ &$-2.7\pm0.21$ \\
& & 75\% &$-2.75\pm0.17$ &$-4.20\pm2.00$ &$-2.08\pm0.20$ \\
& & 100\% &$-3.54\pm0.45$ &$-2.57\pm0.56$ &$-1.91\pm0.04$ \\
\cmidrule[.25pt]{2-6}
%2 cell
& \multirow{3}{*}{2 cell} & 50\% &$-6.70\pm0.62$ &$-3.71\pm0.42$ &$-2.37\pm0.17$ \\
& & 75\%  &$-4.24\pm0.67$ &$-2.23\pm0.23$ &$-1.76\pm0.17$ \\
& & 100\% &$-3.75\pm1.15$ &$-2.58\pm1.46$ &$-2.11\pm0.21$ \\
\bottomrule
\end{tabular}
\end{adjustbox}
\vspace{-10pt}
\end{center}
\end{table}

\subsubsection{Path Length and Shoulder Angle} 
\begin{figure}[!t]
\vspace{8pt}
     \centering
     \includegraphics[width=1
     \columnwidth]{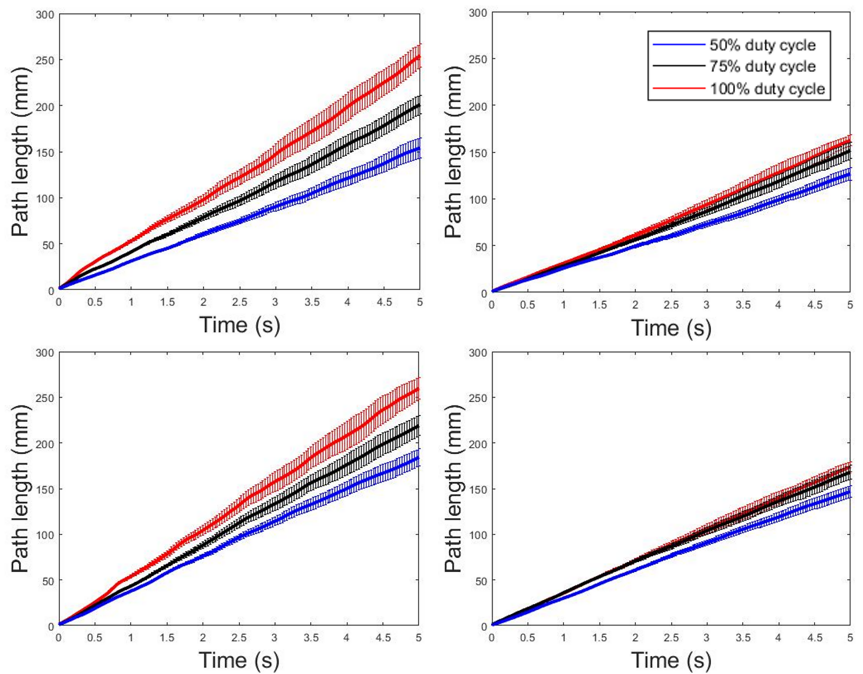}
     \vspace{-6pt}
     \caption{Motion path length of end-effector with 1-cell (left) and 2-cell (right) actuators  at inflation (top) and deflation (bottom) across duty cycles.}
     \label{fig:distance}
     \vspace{-8pt}
\end{figure}

Two-way ANOVAs were performed to identify differences on motion path length across the two types of actuators and the different duty cycles, for each type of motion (abduction and adduction).
For abduction, a significant interaction between number of cells and duty cycles was found ($F[2, 905] =7.59$, $p = 0.001$) supporting that the $\%$ duty cycle affects differently the performance of the two actuators.
Similarly, for adduction, a significant interaction between number of cells and duty cycles was found ($F[2, 905] =3.60$, $p = 0.028$) supporting that the $\%$ duty cycle affects differently the performance of the two actuators.
Motion path profiles are depicted in Fig.~\ref{fig:distance}.

As for the changes in angle at the shoulder joint, two observations can be pointed out. First, the range of motion on the vertical axis is greater with the 1-cell actuator compared to the 2-cell actuator. 
Second, the average maximum angle achieved with the 2-cell actuator at $100\%$ duty cycle is similar to the average maximum angle achieved with the 1-cell actuator at $50\%$ duty cycle, indicating that we can achieve higher elevation of the arm against gravity with the 1-cell actuator (Table \ref{angle}).

\begin{figure}[!t]
\vspace{6pt}
     \centering
     \includegraphics[width=1\columnwidth]{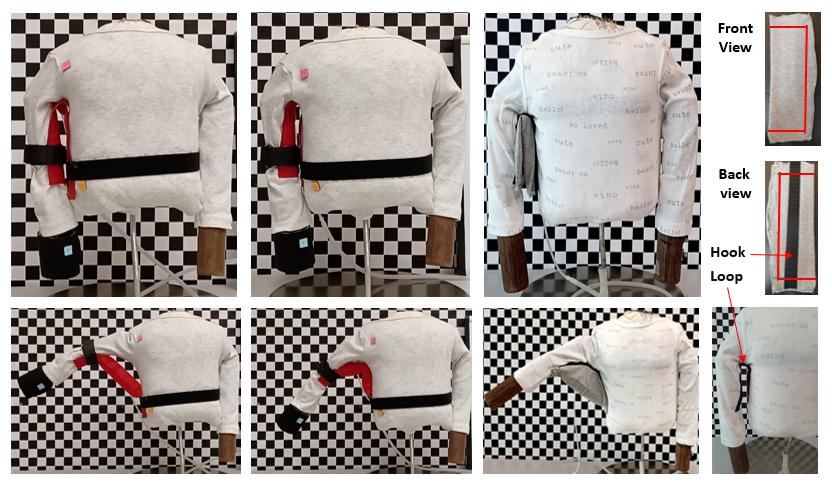}
     \vspace{-6pt}
     \caption{The actuators in deflated (top) and inflated (bottom) states. From left to right: 1-cell, 2-cell, and 1-cell actuator placed within a detachable double-layered pocket. Box in red indicates the housing space of the actuator in the pocket. The hook on the pocket connects with the loop on the wearable.} % in deflated state (left) and at full inflation (right).}
     \label{fig: full_inflation}
     \vspace{-18pt}
\end{figure}

Overall, the aforementioned findings inform next steps into the design and control (more information in \cite{Mucchiani2022closed}) of our wearable device as well as considerations for its use. For example, although the 1-cell actuator achieves greater elevation of the arm and generates greater forces, comes with limitations that can be addressed with the 2-cell actuator. 
Fig.~\ref{fig:xy2} and~Table\ref{CV} present visual and quantitative (Coefficient of Variation [CV]) data that support a greater trial-to-trial variation in force generation and end-effector x,y position across every $\%$ duty cycle for the 1-cell actuator that is not observed with the 2-cell actuator.
In addition, as the 1-cell actuator achieves full inflation and leads to maximum joint angle, it protrudes from the underarm placement. The 2-cell actuator, on the other hand, remains in place throughout the inflation making it more low-profile for our wearable (Fig.~\ref{fig: full_inflation}). 

\begin{table}[!t]
\caption{Shoulder Angle}\label{angle}
\vspace{-12pt}
\begin{center}
%\begin{adjustbox}{width=\columnwidth,center}
\begin{tabular}{c c c}
\toprule
Actuator & Duty cycle & Shoulder Angle (degree) (Mean$\pm$ SD) \\
\midrule
\multirow{3}{*}{1 cell} & 50\% & $41.92\pm1.84$  \\
&  75\% & $53.40\pm1.06$ \\
&  100\% & $63.89\pm1.91$ \\
\midrule
\multirow{3}{*}{2 cell} & 50\% &$34.43\pm0.66$\\
& 75\% & $39.87\pm0.36$ \\
& 100\% & $41.89\pm0.50$  \\
\bottomrule
\end{tabular}
%\end{adjustbox}
\end{center}
\vspace{-12pt}
\end{table}

\begin{table}[!t]
\caption{Coefficient of variation(CV)}\label{CV}
\vspace{-12pt}
\begin{center}
%\begin{adjustbox}{width=\columnwidth,center}
\begin{tabular}{c c c c c}
\toprule
Actuator & Duty cycle & Force & Smoothness & Path length \\
\midrule
\multirow{3}{*}{1 cell} & 50\% & $0.33$ & $0.58$ & $0.59$  \\
&  75\% &  $0.30$ & $0.46$ & $0.58$ \\
&  100\%  & $0.18$ & $0.30$ & $0.57$ \\
\midrule
\multirow{3}{*}{2 cell} & 50\% & $0.09$ & $0.50$ & $0.59$ \\
& 75\% &  $0.04$ & $0.40$ & $0.59$  \\
& 100\% & $0.11$ & $0.44$ & $0.58$  \\
\bottomrule
\end{tabular}
%\end{adjustbox}
\end{center}
\vspace{-12pt}
\end{table}

% Further, post-hoc analysis revealed for the {\bf 1-cell} actuator during $Abduction$, $100$\% duty cycle covers more path than $75$\% ($p<0.001$) and $50$\% ($p<0.001$); and $75$\% duty cycle covers more than $50$\% ($p<0.001$). During $Adduction$, 100\% duty cycle covers more path than 75\% ($p<0.001$) and 50\% ($p=0.02$), but there is no significant difference between 75\% and 50\% duty cycle ($p=0.081$).
% For the {\bf 2-cell} actuator during $Abduction$, there is no significant difference either between $50$\% and $75$\% duty cycle ($p=0.054$) or $75$\% and $100$\% ($p=0.466$), but $100$\% duty cycle covers more path than $50$\% ($p=0.001$). During $Adduction$, there is no significant difference either between $50$\% and $75$\% duty cycle ($p=0.107$) or $75$\% and $100$\% ($p=0.935$), but $100$\% duty cycle covers more path than $50$\% ($p=0.047$).
% %\end{itemize}  

\section{Conclusion and Future Work}
This paper demonstrates the design and evaluation of a family of new fabric-based actuators for the infant shoulder. 
Among 20 initially-proposed multi-cell designs first evaluated in simulation, two actuators (1-cell and 2-cell) were down-selected and tested on a physical infant-sized model.
Overall, the 1-cell actuator seems to generate forces and arm motion that are appropriate for our application; nevertheless, trial-to-trial variation is less for the 2-cell actuator indicating a higher reproducibility.
Although fabric-based actuators may come with limitations such as low force generation \cite{Belforte2014}, for our purpose (i.e. use with infants), we found these forces adequate to lift the arms against gravity. 
Future work will focus on supporting other UE joints, developing closed-loop controllers~\cite{Mucchiani2022closed}, and eventually testing with infants.

\bibliography{ipsita, Kokkoni}
\bibliographystyle{IEEEtran}

\end{document}